\documentclass[lettersize,journal]{IEEEtran}

\usepackage{pifont}%
\newcommand\MYhyperrefoptions{bookmarks=true,bookmarksnumbered=true,
pdfpagemode={UseOutlines},plainpages=false,pdfpagelabels=true,
colorlinks=true,citecolor={black},
pdftitle={Data-Efficient Collaborative Decentralized Thermal-Inertial Odometry},%
pdfsubject={Collaborative localization, drones, space robotics, thermal-inertial odometry},%
pdfauthor={Vincenzo Polizzi, Robert Hewitt,  Javier Hidalgo-Carri\'{o}, Jeff Delaune and Davide Scaramuzza},%
pdfkeywords={Collaborative localization, drones, space robotics, thermal-inertial odometry}}%
\usepackage[\MYhyperrefoptions,pdftex]{hyperref}
\usepackage{adjustbox}
\usepackage{booktabs}
\usepackage{hyperref}
\usepackage{balance}
\usepackage{siunitx}
\usepackage{color}
\usepackage{float}
\usepackage{amssymb}  %
\usepackage{epsfig} %
\usepackage{times} %
\usepackage{mathptmx} %
\usepackage{graphics} %
\usepackage{amsmath,amsfonts}
\usepackage{algorithmic}
\usepackage{algorithm}
\usepackage{array}
\usepackage[caption=false,font=normalsize,labelfont=sf,textfont=sf]{subfig}
\usepackage{textcomp}
\usepackage{stfloats}
\usepackage{url}
\usepackage{verbatim}
\usepackage{graphicx}
\usepackage{cite}
\usepackage{orcidlink}

\newcommand{\yesmark}{\textcolor{green}{\ding{51}}}%
\newcommand{\nomark}{\textcolor{red}{\ding{55}}}%

\input{sections/macros}

\usepackage[capitalize]{cleveref}
\crefname{section}{Sec.}{Secs.}
\Crefname{section}{Section}{Sections}
\Crefname{table}{Table}{Tables}
\crefname{table}{Tab.}{Tabs.}

\title{
Data-Efficient Collaborative Decentralized Thermal-Inertial Odometry
}

\author{Vincenzo Polizzi$^{\orcidlink{0000-0003-4900-7898}}$, Robert Hewitt$^{\orcidlink{0000-0003-0895-6747}}$,  Javier Hidalgo-Carri\'{o}$^{\orcidlink{0000-0001-6709-9285}}$, Jeff Delaune$^{\orcidlink{0000-0003-1509-4401}}$ and Davide Scaramuzza$^{\orcidlink{0000-0002-3831-6778}}$ 

\thanks{Manuscript received 24 February 2022; accepted 4 July 2022. Date of publication 28 July 2022; date of current version 9 August 2022. This letter was recommended for publication by Associate Editor P. Pounds and Editor P. Castillo upon evaluation of the reviewers' comments. The research was funded by the Combat Capabilities Development Command Soldier Center and Army Research Laboratory. This research was carried out at the Jet Propulsion Laboratory, California Institute of Technology, and was sponsored by the JPL Visiting Student Research Program (JVSRP) and the National Aeronautics and Space Administration (80NM0018D0004). \emph{(Corresponding author: Polizzi
Vincenzo.)}}
\thanks{Vincenzo Polizzi is with the Robotics and Perception Group, University of
Zurich, 8006 Zürich, Switzerland, and also with the Jet Propulsion Laboratory, California Institute of Technology, Pasadena, CA 91125 USA (e-mail:
vincenzo.polizzi@ieee.org).}%
\thanks{Javier Hidalgo-Carrió and Davide Scaramuzza are with the Robotics and
Perception Group, University of Zurich, 8006 Zürich, Switzerland (e-mail:
jhidalgocarrio@ifi.uzh.ch; davide.scaramuzza@ieee.org)}%
\thanks{Robert Hewitt and Jeff Delaune are with Jet Propulsion Laboratory, California Institute of Technology, Pasadena, CA 91125 USA (e-mail: robert.
a.hewitt@jpl.nasa.gov; jeff.h.delaune@jpl.nasa.gov).}
\thanks{Digital Object Identifier 10.1109/LRA.2022.3194675}
}

\begin{document}

\maketitle

\begin{abstract}
We propose a system solution to achieve data-efficient, decentralized state estimation for a team of flying robots using thermal images and inertial measurements. \changes{Each robot can fly independently, and exchange data when possible to refine its state estimate.}
Our system front-end applies an online photometric calibration to refine the
thermal images so as to enhance feature tracking and place recognition. Our
system back-end uses a covariance-intersection fusion strategy to neglect
the cross-correlation between agents so as to lower memory usage and
computational cost. The communication pipeline uses Vector of Locally
Aggregated Descriptors (VLAD) to construct a request-response policy that
requires low bandwidth usage. We test our collaborative method on both
synthetic and real-world data. Our results show that the proposed method
improves by up to~\SI{46}{\percent} trajectory estimation with respect to an
individual-agent approach, while reducing up to~\SI{89}{\percent} the
communication exchange. Datasets and code are released to the public,
extending the already-public JPL xVIO library.

\end{abstract}


\section*{Supplementary Material} \label{sec:SupplementaryMaterial}
\hspace{-0.5cm} For code and datasets please visit: \url{https://rpg.ifi.uzh.ch/xctio}
\section{Introduction}
\label{sec:intro}
\vspace{-0.3ex}

\IEEEPARstart{W}{ith} the successful deployment of the Ingenuity Mars Helicopter and the
Perseverance rover on
Mars,\footnote{[Online]. Available: \url{https://www.nytimes.com/2022/02/15/science/mars-nasa-perseverance.html}}
multi-agent collaborative tasks are now being carried out on another world for
the first time. Ingenuity performed 19 flights during its first year of
operations, many of which were done in direct support of Perseverance’s
exploration of Jezero crater. This continues a new era of exploration for NASA,
started by the MarCO mission in 2018~\cite{MarCO}, which saw a pair of
co-dependent cubesats investigate Mars for the first time. Next, NASA’s CADRE
mission plans to send a group of small rovers to cooperatively explore the Lunar
surface in 2024 as part of its Commercial Lunar Payload Services
program~\cite{CADRE}. 

This trend towards smaller, cooperative agents is reflected by the more
dangerous terrain NASA wishes to explore in future missions to the Moon, Mars
and beyond. These areas include Recurring Slope Lineae on Mars and lava tubes on
the Moon and Mars. Multiple small agents can be released from larger rovers or
deployed from landers to explore areas that are too risky for a single agent to
explore. It’s also possible for multiple agents to explore large regions
quickly, and deviate from the primary mission task without risk of jeopardizing
the primary mission timeline. 

\begin{figure}[t!]
    \centering
    \includegraphics[width=\linewidth, trim=3.5 5.0 4.0 0.2, clip]{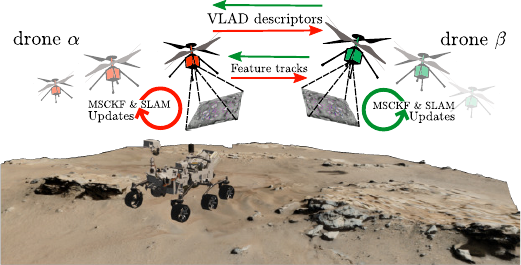} \hfill
    \vspace{-3ex}
    \caption{
        Schematic representation of the proposed method with two flying robots in a Mars scenario.
    }
    \label{fig:eyecatcher}
    \vspace{-4ex}
\end{figure}

The ability for agents to localize themselves in an unknown environment with no
prior infrastructure (e.g., GPS) is a key capability for autonomous operations
of multiple agents (see~\cref{fig:eyecatcher}).  Visual-Inertial Odometry (VIO)~\cite{Zhang19Springer,Huang19icra}
is one navigation alternative that the Ingenuity Mars Helicopter utilizes to
estimate its pose during flights~\cite{bayard2019vision}.  Additionally,
pre-cursors to the CADRE mission, such as the Autonomous-PUFFER research task
\cite{APUFFER}, have demonstrated the ability of multiple agents to exchange
information while performing localization to improve their overall estimate by
working together while exploring their work area. 

At the same time, there has also been growing interest and work to incorporate
new sensing inputs to autonomous navigation algorithms that do not rely on day
light conditions and can be used to operate these algorithms in extreme
locations such as lava tubes that are deprived of external lighting. One of the
sensor technologies being investigated for this purpose are thermal cameras,
which measure thermal radiation from all objects in their field of
view~\cite{SurveyThermalCameras}. These types of cameras allow autonomous agents
to operate in all types of lighting conditions and require no external lighting
source. 

This paper aims to incorporate these two separate threads (multi-agent
localization and thermal sensing) into a single system based around multiple
aerial vehicles to investigate the benefits that such a system could bring to future
missions. We present a collaborative VIO system architecture that allows agents
to operate in a decentralized way while still making use of information that is
received from other agents. In addition, we have incorporated recent
advancements in thermal image calibration into our system to improve VIO
performance on thermal imagery. Our contributions are:
\vspace{-0ex}
\begin{itemize}
\item The development of a scalable\changes{, from 1 to N agents,} Collaborative Thermal-Inertial Odometry using a photometric calibration algorithm.
\changes{\item The incorporation of Binary VLAD descriptors in a thermal-inertial odometry solution.}
\item A scalable and light communication approach over a decentralized \changes{flying robot} system.
\item Code and datasets are released as part of the already public
    JPL xVIO library\footnote{[Online]. Available: \url{https://github.com/jpl-x/x}}~\cite{delaune20xvio}.
\end{itemize}

\section{Related Work}
\label{sec:related}
State estimation for multi-agent systems is a well-known and established problem which has recently gained attention. The solutions can be divided into centralized and decentralized approaches. The former relies on a central unit acting as a server, fusing incoming information from all the agents~\cite{Schmuck17icra,
Schmuck21ismar}. \changes{The latter directly fuses data onboard  requiring efficient fusion processes and low data exchange~\cite{DecentralisedSLAMlowband}. While a centralized version is easier to implement, it relies on a central entity to always be reachable, never fail, and scale with the number of agents, limiting the range of exploration. This work focuses on the decentralized approach to be robust and scalable~\cite{gautam12iciis}.}

Decentralized \changes{Simultaneous Localization and Mapping (SLAM)} faces the problem of exploiting communication. Indirect measurements through the registration of observations are the preferable approach in many scenarios~\cite{Dong15icra}. Place recognition has emerged as a compact and data-efficient solution~\cite{Arandjelovic16cvpr}. The work in~\cite{Cieslewski18icra} uses \changes{Vector of Locally Aggregated Descriptors (VLADs)} generated by the NetVLAD \changes{\cite{ArandjelovicNetVLAD16}} to reduce the communication bandwidth. Instead of sending information to all the other agents, the agent sends data only to the robot assigned with a specific descriptor. Although this strategy lowers the used bandwidth, it limits the system's scalability due to the a priori assignment of a centroid of the clusterized feature space. We overcome this problem and generate binary VLADs from the feature descriptors extracted from the frames.

The evaluation of graph-based optimization vs. filter-based approaches for structure from motion has been rigorously analyzed in~\cite{Strasdat12jivc}. The conclusion is that optimization approaches outperform filtering in terms of accuracy per unit of computing time. Visual SLAM improves accuracy by increasing
the number of features while having minor effects when increasing the number of keyframes.  However, the analysis was done with a standard \changes{Extended Kalman filter (EKF)} formulation, which models the features in the vector state, and the computational time grows quadratically with the number of observations.  Other approaches such \changes{as} the \changes{Multi-State Constrained Kalman Filter (MSCKF)}~\cite{Mourikis07icra} \changes{express} the constraints without including the 3D feature position in the filter state vector, resulting in computational complexity linear in the number of features. The work in~\cite{Li13ijrr} showed that MSCKF competes with graph-based optimization approaches in terms of accuracy and computational cost.  We focus our method on a filter-based approach using the MSCKF formulation of the JPL xVIO library, targeting onboard rotorcraft flight applications. We extend the xVIO back-end to fuse multi-agent data, developing a collaborative formulation of the MSCKF. We use the Covariance Intersection (CI) algorithm~\cite{609105} in order to avoid \changes{the filter being overconfident} in the resulting updates. The CI overestimates the resulting covariance after fusing \changes{the measurements} coming from several agents, ensuring filter consistency. Our back-end is similar to the research in~\cite{Zhu21icra} and~\cite{Zhu21iros}.

Thermal-inertial solutions have recently received considerable attention for odometry systems in challenging scenarios. The technology was not widely used due to \changes{its} cost and poor resolution. However, recent advances in thermal imaging have made thermal cameras more affordable, with \changes{a} smaller footprint, higher quality and modest power consumption enabling the portability to robotics applications~\cite{SurveyThermalCameras}.
The work by Delaune et al.~\cite{Delaune19iros}, and Khattak et al.~\cite{khattak19icra}, have shown the use of raw thermal data to navigate in dark scenarios where conventional visual cameras fail. Despite the outstanding results obtained by these works, thermal-infrared cameras \changes{suffer from} photometric inconsistencies due to the thermography generation process. As a result, a standard computer vision program that relies on the photometric consistency assumption might fail to deliver reliable perception data in a sequence of frames.  In this direction, the work by Das et al.~\cite{Das2021}, develops a photometric model of Thermal Infrared (TIR) cameras to attenuate such disturbances, obtaining photometric consistency over time and space.  However, place recognition with thermal imaging remains an open challenge, and its application in collaborative systems has not been investigated.

\begin{table}[tb!]
\centering
\caption{\changes{Overview of the strengths of state-of-the-art visual-inertial odometry frameworks.}
}
\begin{adjustbox}{max width=1.0\linewidth}
\setlength{\tabcolsep}{6pt}
\begin{tabular}{lccccl}
 & \textbf{Thermal} & \textbf{Collaborative} & \textbf{Decentralized} & \textbf{Data-Efficient} & \textbf{Remarks}\\
\midrule
Schmuck et al.~\cite{Schmuck17icra, Schmuck21ismar} & \nomark & \yesmark & \nomark & \nomark & Centralized with loop closure\\
Cieslewski et al.~\cite{Cieslewski18icra} & \nomark & \yesmark & \yesmark & \yesmark & Graph-based NetVLAD\\
Zhu et al.~\cite{Zhu21icra, Zhu21iros} & \nomark & \yesmark & \yesmark & \nomark & SLAM and Kalman features\\
Delaune et al.~\cite{Delaune19iros} & \yesmark & \nomark & \nomark & \nomark & Thermal intertial odometry\\
\textbf{This work} & \yesmark & \yesmark & \yesmark & \yesmark & Filter-based Thermal VLAD \\
\bottomrule
\end{tabular}
\end{adjustbox}
\vspace{-1ex}
\label{tab:soa:methods}
\vspace{-4ex}
\end{table}

~\cref{tab:soa:methods} summarizes the related work on collaborative visual odometry systems. Our proposal is a fully decentralized thermal inertial odometry (TIO) that uses place recognition with thermal imaging and benefits from a data-efficient collaborative strategy. To the best of our knowledge, this path has not been explored.

\section{Methodology}
\label{sec:method}
We have a set $\mathbb{U}$ of $U$ \changes{Unmanned Aerial Vehicles (UAVs)}, equipped with a monocular thermal-infrared camera and IMU. Each drone can fly independently by running the JPL xVIO~\cite{delaune20xvio}. \changes{At start-up, each drone is initialized with the same inertial reference frame.}
The system extracts \changes{Features from accelerated segment test (FAST)} corners~\cite{11744023_34} and uses the pyramidal implementation of the Lucas-Kanade tracker~\cite{Bouguet1999PyramidalIO} to build tracks out of the extracted features. We associate an \changes{Oriented FAST and Rotated BRIEF (ORB)} descriptor~\cite{6126544} to each track to perform track matching between the current UAV tracks and the ones received from the other agents. In xVIO there are three kinds of tracks:
\begin{itemize}
    \item \textbf{Opportunistic}: concatenation of the Lucas-Kanade tracker results over a sequence of frames. 
    \item \textbf{SLAM}: tracks \changes{corresponding} to the feature points saved in the state vector. For these points we know the inverse depth parameterization.
    \item \textbf{MSCKF}: opportunistic tracks that have a size between two and $M$ frames and fulfill the requirements to perform an MSCKF update, that is the baseline between two features in the track is higher than a certain threshold. 
\end{itemize}

\begin{figure}[t!]
    \centering
    \includegraphics[width=0.45\textwidth]{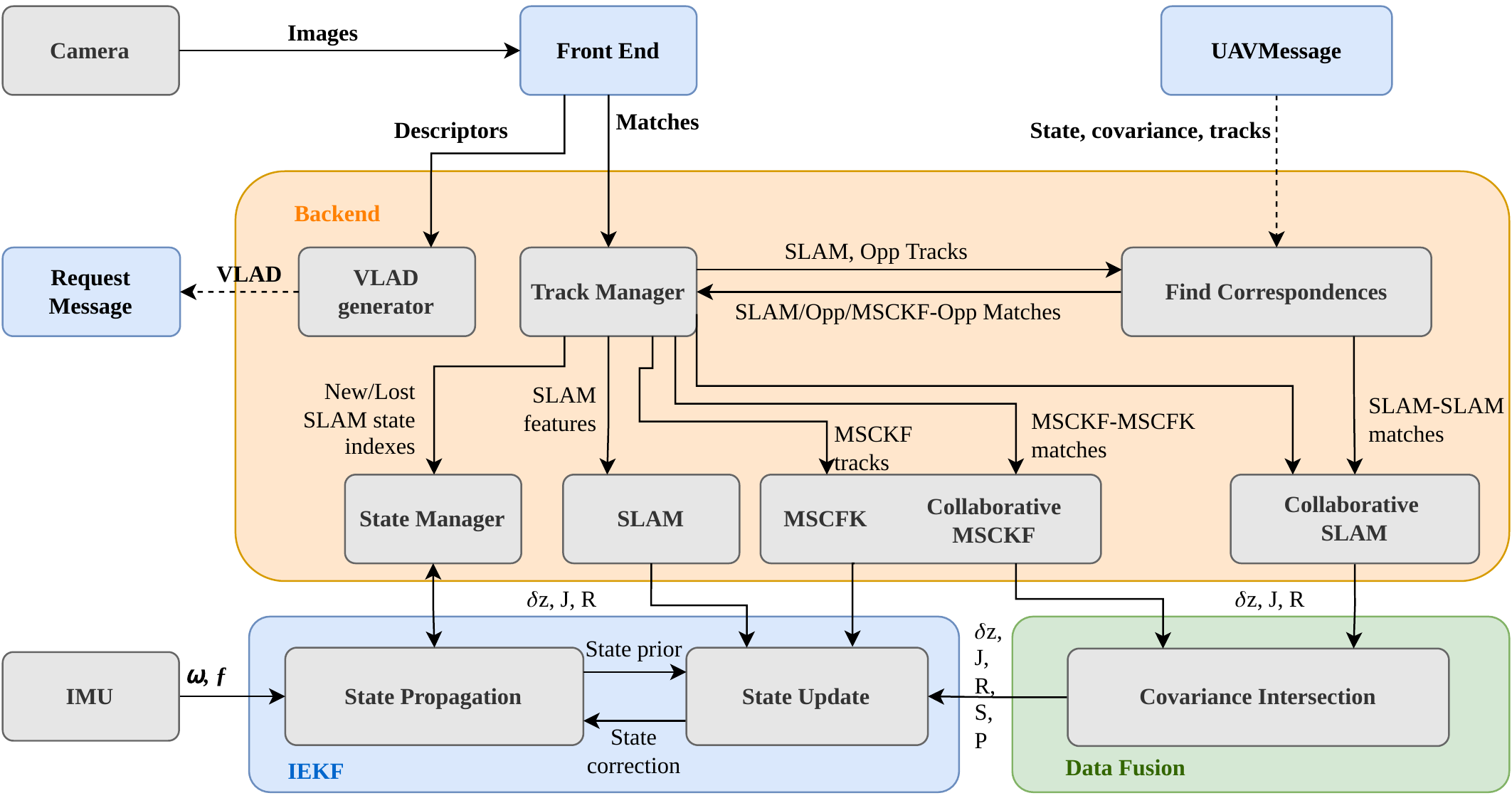}
    \caption{Block diagram of our proposed method. Our solution can receive messages from the other agents and find SLAM-SLAM  and  MSKCF-MSCKF  matches  between  the received data and the local tracks.}
    \label{fig:method}
    \vspace{-2ex}
\end{figure}

An MSCKF track contains a wide baseline between the tracked points, hence a match between opportunistic tracks of two different UAVs fulfill the requirements to define a single MSCKF track. For this reason, hereafter, we will talk only about MSCKF and SLAM tracks. \figref{fig:method} depicts a diagram of our method implemented in the xVIO library. The front-end provides feature point descriptors and generates a VLAD descriptor contained in the request message. The Track Manager unit receives the  matches, stores them and forms tracks. The tracks are employed to perform MSCKF and SLAM updates: when a message is received, the back-end finds correspondences between the current tracks and the received ones. The matches generated from the correspondences are then processed as SLAM-SLAM matches updates, or stored and processed afterward as collaborative MSCKF updates.

Details of the the EKF updates, the state and covariance propagation, the thermal front-end and the communication pipeline are described in the following.

\subsection{Thermal Front-end}
\begin{figure}[t]
  \centering
  \includegraphics[width=0.493\linewidth]{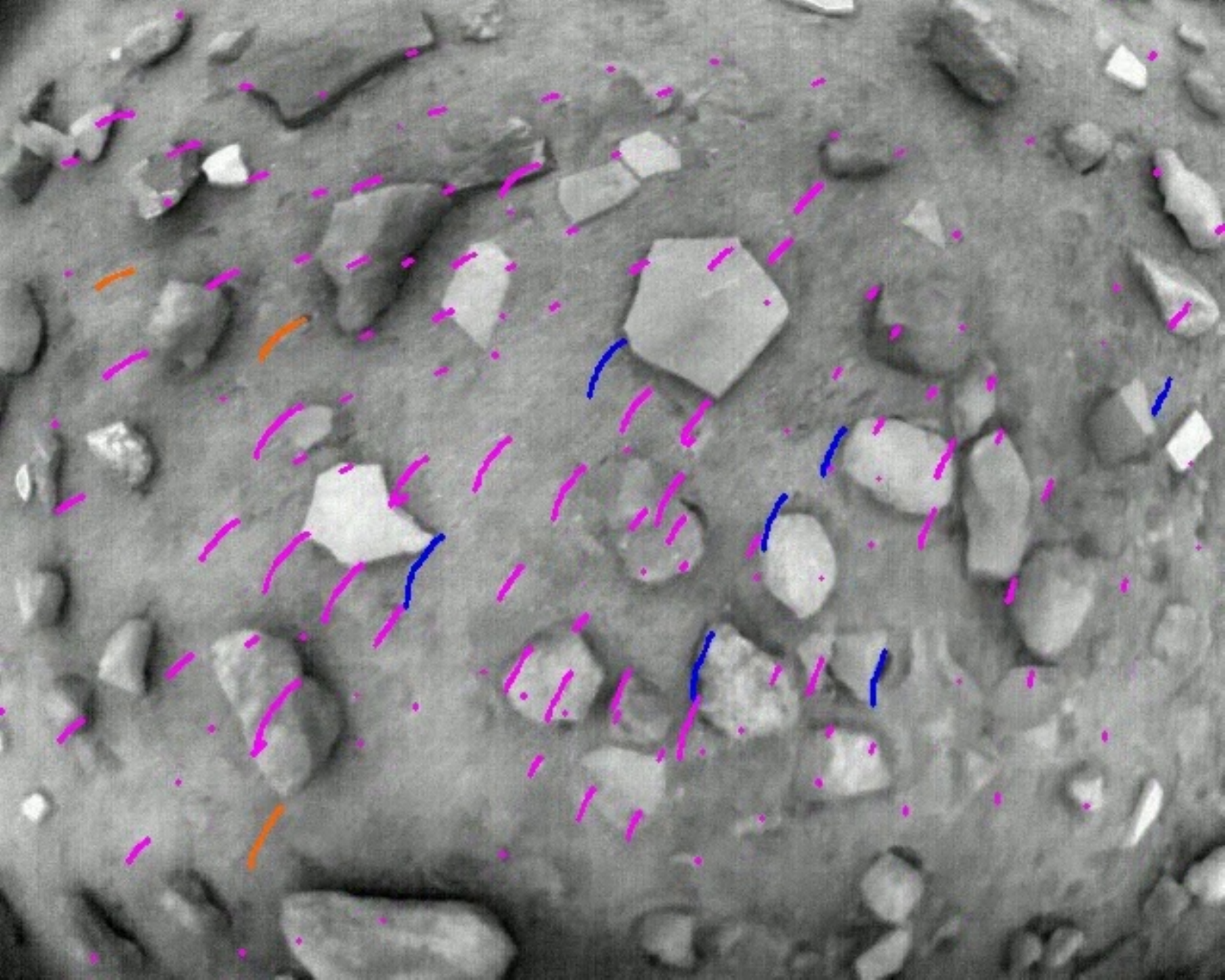}\hfill
  \includegraphics[width=0.493\linewidth]{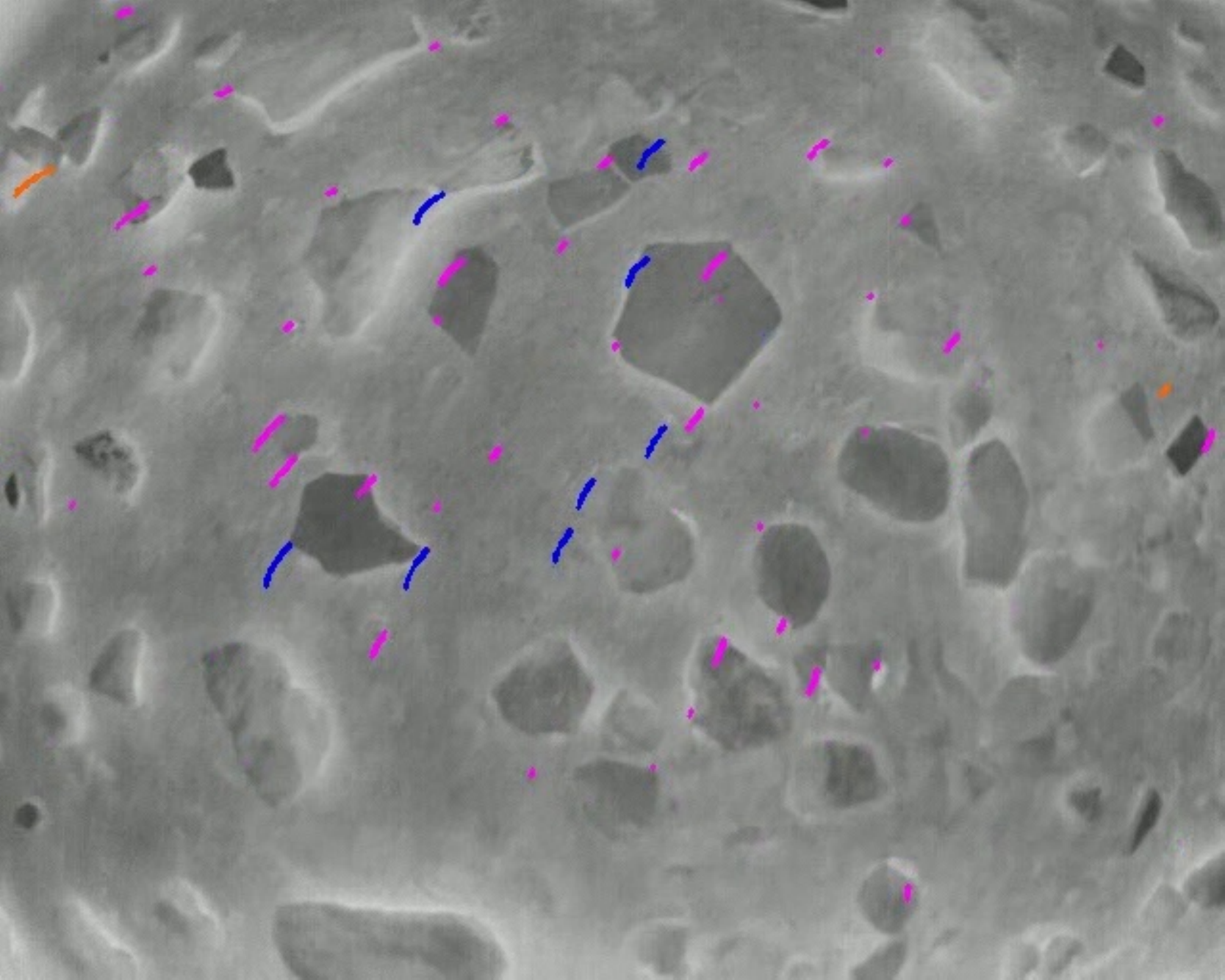}
  \\[\smallskipamount]
  \caption{Tracks representation in (a) calibrated and (b) uncalibrated thermal image. Blue SLAM tracks, orange MSCKF tracks, purple Opportunistic tracks.}
 \label{fig:tracks_draw}
\end{figure}

\begin{figure}[t]
  \centering
  \includegraphics[width=0.95\linewidth]{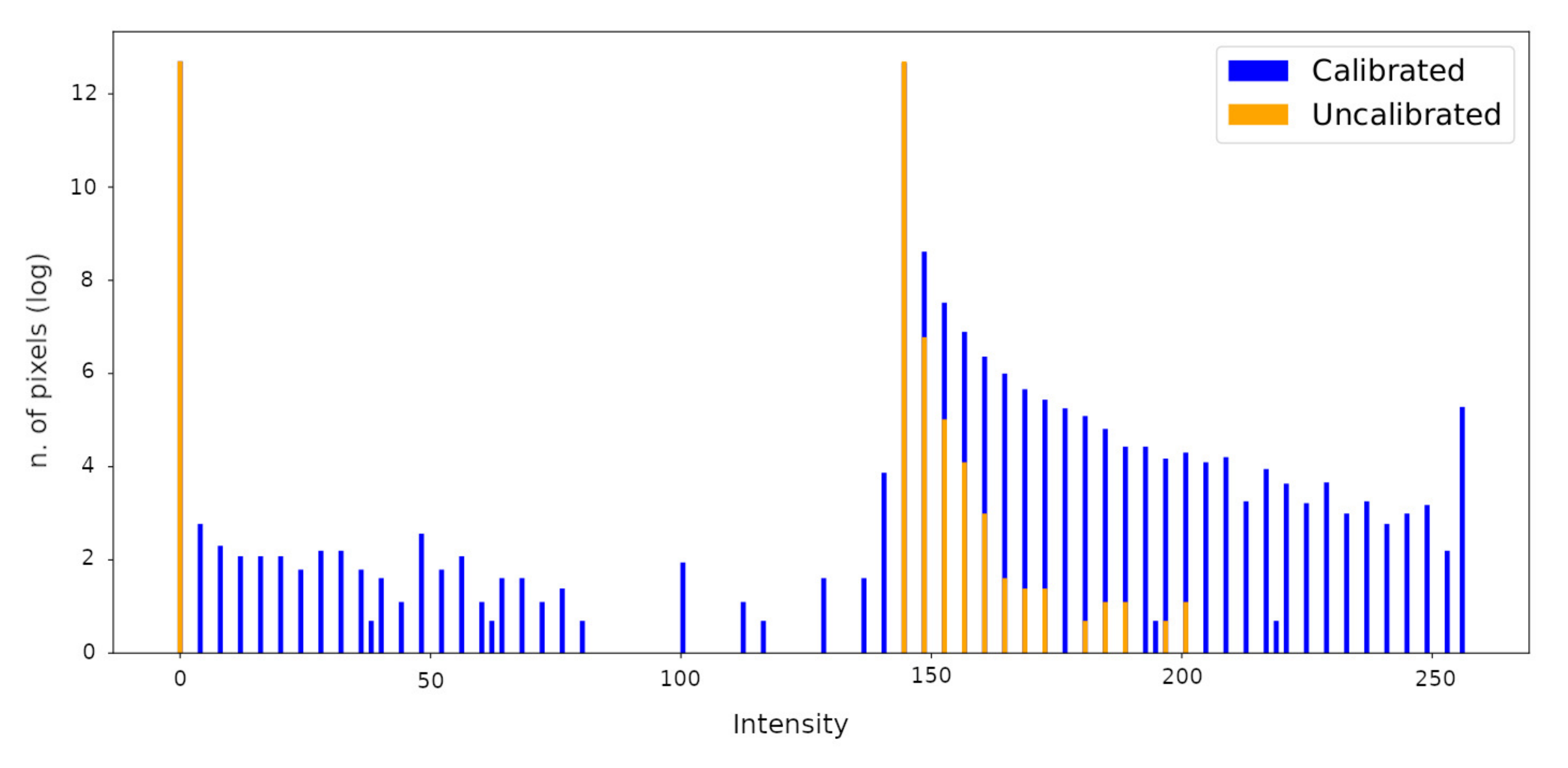}
  \vspace{-1ex}
  \caption{\changes{The histogram of the corner response of the calibrated and uncalibrated data, shows that the calibrated images have a higher Harris cornerness response.}}
  \label{fig:corner_histogram}
  \vspace{-1ex}
\end{figure}

Thermal images show low contrast and bring photometric inconsistency over a sequence of frames. To deal with these problems, we employed the work of Das et~al.~\cite{Das2021} that attenuates the low-frequency non-uniformness and photometric inconsistency. \figref{fig:corner_histogram} shows that calibrated images return a higher Harris cornerness response than the uncalibrated ones, and hence make it possible to detect and track stronger corners. 
FAST corner detector compares the intensity value of a central pixel with a ring of pixels around it. A corner is detected if the intensity of $N$ contiguous pixels in the ring is higher or lower than the central one by a threshold $t$. With calibrated thermal frames, the gradient of the image is high, and hence the value of $t$ can be set to detect the most distinctive corners.
We counted the number of tracks and evaluated the feature life for both calibrated and uncalibrated frames. As shown in \figref{fig:tracks_draw}, applying spatial and temporal parameters calibration results in approximately four times the number of FAST corners being successfully detected and tracked with the Lucas-Kanade tracker than on uncalibrated images, with average feature track length on calibrated images also being increased. Namely, it is harder to lose the tracked corners since the image has higher contrast than the original thermal frame. This result implies that since four times more tracks are detected, we can potentially have four times more track correspondences between the robots, hence more collaborative updates for the EKF.


\subsection{State definition}
\label{sec:state-definition}
In the xVIO back-end, the state vector $\state$ \eqref{eq:state-vec-ivr}, can be divided into the states concerning the IMU $\sxi$ and the visual part $\xv$.

The first term, $\sxi \in \mathbb{R}^{16}$, defined in \eqref{eq:state-vec-i} includes the position, velocity and orientation quaternion of the IMU frame $\left\{\mathit{i}\right\}$ with respect to the world frame $\left\{\mathit{w}\right\}$, the gyroscope and the accelerometer biases, $\boldsymbol{b}_{g}$ and $\boldsymbol{b}_{a}$ respectively.

The second term, $\xv \in \mathbb{R}^{7M+3N}$, defined in \eqref{eq:state-vec-s} is split into the sliding window $\xs$, that has the positions $\left\{\boldsymbol{p}_w^{c_i}\right\}_{i}$ and the orientations $\left\{\boldsymbol{q}_{w}^{c_i}\right\}_{i}$ of the camera frame of the last $M$ image time instances, and the feature states $\xf$ \eqref{eq:state-vec-f}, that contains the inverse depth parametrization of the landmarks $f_i$ with $ i=1,...,N$.
\begin{equation}
\boldsymbol{x} = \begin{bmatrix}{\sxi}^T &
{\xv}^{T}\end{bmatrix}^{T}
\label{eq:state-vec-ivr}
\end{equation}
\begin{equation}
\sxi =
\begin{bmatrix}
{\boldsymbol{p}_w^i}^T &
{\boldsymbol{v}_w^i}^T &
{\boldsymbol{q}_{w}^{i}}^T &
{\boldsymbol{b}_{g}}^{T} &
{\boldsymbol{b}_{a}}^{T}
\end{bmatrix}^{T}
\label{eq:state-vec-i}
\end{equation}
\begin{equation}
\xs =
{\begin{bmatrix}
{\pos{w}{c_1}}^T & ... & {\pos{w}{c_M}}^T &
{\quat{w}{c_1}}^T & ... & {\quat{w}{c_M}}^T
\end{bmatrix}}^T
\label{eq:state-vec-s}
\end{equation}
\begin{equation}
\xf =
{\begin{bmatrix}
{\boldsymbol{f}_{1}}^T & ... & {\boldsymbol{f}_{N}}^T
\end{bmatrix}}^T
\label{eq:state-vec-f}
\end{equation}
The error state vector $\deltastate$ can also be divided as the state vector. The error state is defined for the position vectors as the difference between the true and estimated position, $\pos{w}{i}-\estpos{w}{i}=\delta{\pos{w}{i}} \in \mathbb{R}^{3}$, and similarly for the velocity and inertial bias errors. Differently, for the quaternions, the error vector is given by $\quat{w}{i}=\estquat{w}{i}	\otimes \delta{\quat{w}{i}} \in \mathbb{R}^4$, where $\otimes$ is the quaternion product. For a minimal representation of the error quaternion, we use the small angle approximation, namely $\delta\boldsymbol{q} \simeq \begin{bmatrix}
1 &
\frac{1}{2}{\delta\boldsymbol{\theta}}^{T}\end{bmatrix}^{T}$, hence $\delta\boldsymbol{\theta} \in \mathbb{R}^{3}$ represents the quaternion error in a lower dimension space.

The measurement model of 3D landmark $\wpf$ in the world frame $\{w\}$, observed by the camera $\{c_i\}$, is the normalized projection on the unit focal length camera plane:
\begin{equation}
^i\boldsymbol{z}_j =
\frac{1}{^{c_i}z_j}
\begin{bmatrix}^{c_i}x_j\\^{c_i}y_j\end{bmatrix}
+ {^i\boldsymbol{n}_j}
\label{eq:vision-measurement}
\end{equation}
where
\begin{align}
^{c_i}\boldsymbol{p}_j &= \begin{bmatrix}
^{c_i}x_j & ^{c_i}y_j & ^{c_i}z_j \end{bmatrix}^T
\label{eq:feature-cart-coord}\\
&= \boldsymbol{C}(\boldsymbol{q}_w^{c_i})({^w\boldsymbol{p}_j}
- \boldsymbol{p}_w^{c_i})
\label{eq:feature-frame-tf1}
\end{align}
and $^i\boldsymbol{n}_j$ is a zero-mean white Gaussian measurement noise with
covariance matrix $^i\boldsymbol{R}_j = \sigma_V^2\boldsymbol{I}_2$. We assume the standard deviation $\sigma_V$ is uniform over the image and it depends on the performance of the visual front-end. $\boldsymbol{C}(\boldsymbol{q}_a^{b})$ is the rotation matrix associated with the quaternion $\boldsymbol{q}_a^{b}$ such that ${}^bx=\boldsymbol{C}(\boldsymbol{q}_a^{b}){}^ax$.

\subsection{MSCKF-MSCKF update}

\begin{figure}[t!]
    \centering
    \includegraphics[width=0.40\textwidth]{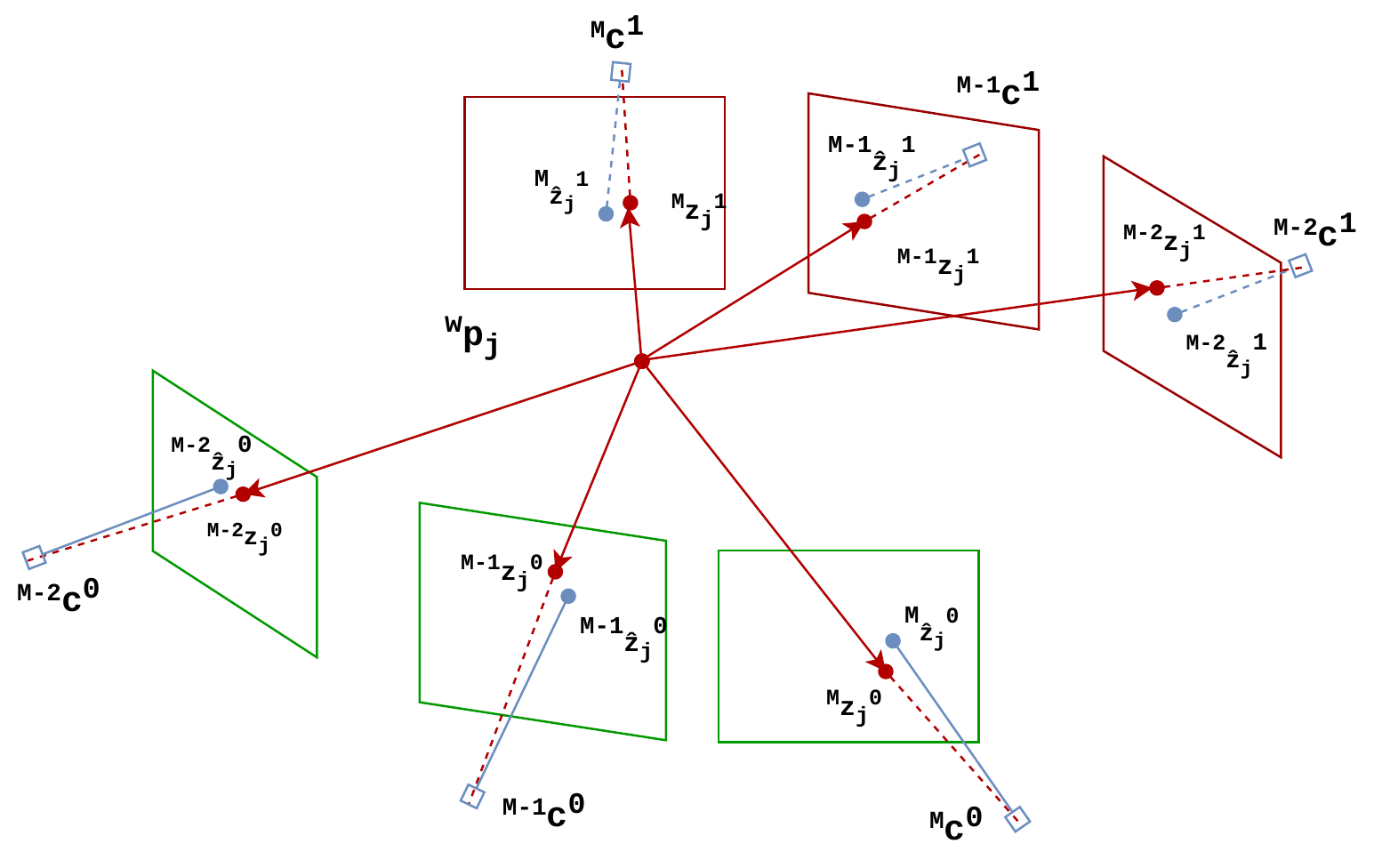}
    
    \caption{Collaborative MSCKF representation. $^ic^0$ and $^ic^1$ are the cameras of two different agents $0$ (green) and $1$ (red) respectively, that are tracking the same landmark $\wpf$.}
    \label{fig:collab_msckf}
    \vspace{-2ex}
\end{figure}

We used the same approach proposed by Zhu~\cite{Zhu21icra}. Here we want to give a detailed overview of the method, and how this is applied to the \emph{xVIO} stack.

The \emph{xVIO} back-end performs MSCKF~\cite{Mourikis07icra} updates. MSCFK minimizes the residual between tracked feature points and the back-projection on the camera plane of the triangulated landmark. \figref{fig:collab_msckf} represents how the MSCKF residuals can be interpreted in a collaborative setup. The tracked landmark $\wpf$ is triangulated by using all the measurements of robots' cameras, $^ic^0$ and $^ic^1$. Then the resulting 3D point is projected into the camera planes that observed the feature.

We express the measurement model $\residualp{i}{j}{l} \in \mathbb{R}^2$ of the observed feature ${^i\boldsymbol{\hat{z}}_j^l}$ in the camera frame $i \in \mathbb{I} \subseteq \{1,...,M\}$ of the robot $l \in \mathbb{U}$ using the model in  \eqref{eq:feature-frame-tf1} as a nonlinear function of the state $\state$ and the landmark $\wpf$ \eqref{eq:nonlinear_model}, such that the measurement innovation $\deltaresidualp{i}{j}{l}=\residualp{i}{j}{l}-{^i\boldsymbol{\hat{z}}_j^l}$ can be linearized as in \eqref{eq:linear_model_msckf0}.
\begin{equation}
\residualp{i}{j}{l} = \boldsymbol{h}(\state^l,\wpf)+{^i\boldsymbol{n}_j^l}
\label{eq:nonlinear_model}
\end{equation}
\begin{equation}
\deltaresidualp{i}{j}{l} \simeq \jacob{(i,j)}{\state}{l} \deltastate^{l}+
\jacob{(i,j)}{\boldsymbol{p}}{l} {^w\delta\boldsymbol{p}}_j +
{^i\boldsymbol{n}_j^l}
\label{eq:linear_model_msckf0}
\end{equation}
By stacking together the residuals for all the observations of the landmark $\deltawpf$ for the robot ${l}$ we write:
\begin{equation}
\deltaresidualp{}{j}{l} \simeq 
\jacob{j}{\state}{l} \deltastate^{l}+
\jacob{j}{\boldsymbol{p}}{l} {^w\delta\boldsymbol{p}}_j +
{\boldsymbol{n}_j^l}
\label{eq:linear_model_msckf}
\end{equation}
Since ${}^w{\delta\boldsymbol{p}_j}$ is neither part of the state, nor can be included in the noise, we need to get rid of it to perform an EKF update. With this purpose, we multiply on each side by the left nullspace ${\boldsymbol{A}_{j}}$ of $\jacob{j}{\boldsymbol{p}}{l}$. By doing so we split the system into two subsystems, one $\deltaresidualp{}{0_j}{l}$ that is a function of $\deltastate^{l}$ only and one $\deltaresidualbarp{}{0_j}{l}$ that also depends on $\deltawpf$ \eqref{eq:nullspace_sys}.
\begin{equation}
\begin{bmatrix}
\deltaresidualp{}{0_j}{l} \\
\deltaresidualbarp{}{0_j}{l}
\end{bmatrix}
= 
\begin{bmatrix}
\jacob{j}{0_{\state}}{l}\\
\jacobbar{j}{0_{\state}}{l}
\end{bmatrix}
\deltastate^{l}  + 
\begin{bmatrix}
\jacob{j}{0_{\boldsymbol{p}}}{l} \\
0
\end{bmatrix}
{^w\delta\boldsymbol{p}}_j + 
\begin{bmatrix}
\boldsymbol{n}_{j}^{l}\\
\bar{\boldsymbol{n}_{j}}^{l}
\end{bmatrix}    
\label{eq:nullspace_sys}
\end{equation}
The equations \eqref{eq:linear_model_msckf0}-\eqref{eq:nullspace_sys}, describe the standard procedure to perform an MSCKF update for the agent $l$. In particular in a single UAV update, we feed the EKF with the bottom part of the system \eqref{eq:nullspace_sys}. To perform the collaborative MSCKF update, we need to write a residual that depends on the states of all the agents that observed the same landmark $\wpf$. Writing such a residual is possible by stacking together all the $\deltaresidualp{}{0_j}{l}$ in \eqref{eq:nullspace_sys} with $l \in \mathbb{L}$ where $\mathbb{L}=\{l_0,...l_L\}$ is the set of the $L$ robots that observe $\wpf$ \eqref{eq:collab_msckf}.
\begin{equation}\nonumber
\begin{bmatrix}
\deltaresidualbarp{}{0_j}{l_0} \\
\vdots \\
\deltaresidualbarp{}{0_j}{l_L}
\end{bmatrix}
=
\boldsymbol{D}(
\begin{bmatrix}
\jacob{j}{0_{\state}}{l_0}\\
\vdots \\
\jacob{j}{0_{\state}}{l_L}
\end{bmatrix}
)
\begin{bmatrix}
\deltastate^{l_0} \\
\vdots \\
\deltastate^{l_L}
\end{bmatrix}  + 
\begin{bmatrix}
\jacob{j}{0_{\boldsymbol{p}}}{l_0} \\
\vdots \\
\jacob{j}{0_{\boldsymbol{p}}}{l_L}
\end{bmatrix}
{^w\delta\boldsymbol{p}}_j + 
\begin{bmatrix}
\boldsymbol{n}_{j}^{l_0}\\
\vdots \\
\boldsymbol{n}_{j}^{l_L}
\end{bmatrix}
\end{equation}
\begin{equation}
\Rightarrow
\deltaresidualbarp{}{0_j}{}
= 
\jacob{j}{0_{\state}}{}
\deltastate + 
\jacob{j}{0_{\boldsymbol{p}}}{}
{^w\delta\boldsymbol{p}}_j + 
\boldsymbol{n}_{j}^{}
\label{eq:collab_msckf}
\end{equation}
Where $\boldsymbol{D}(\cdot)$ represents the matrix having a diagonal that corresponds to the elements of the input vector.

\begin{figure}[t!]
    \centering
    \includegraphics[width=0.40\textwidth]{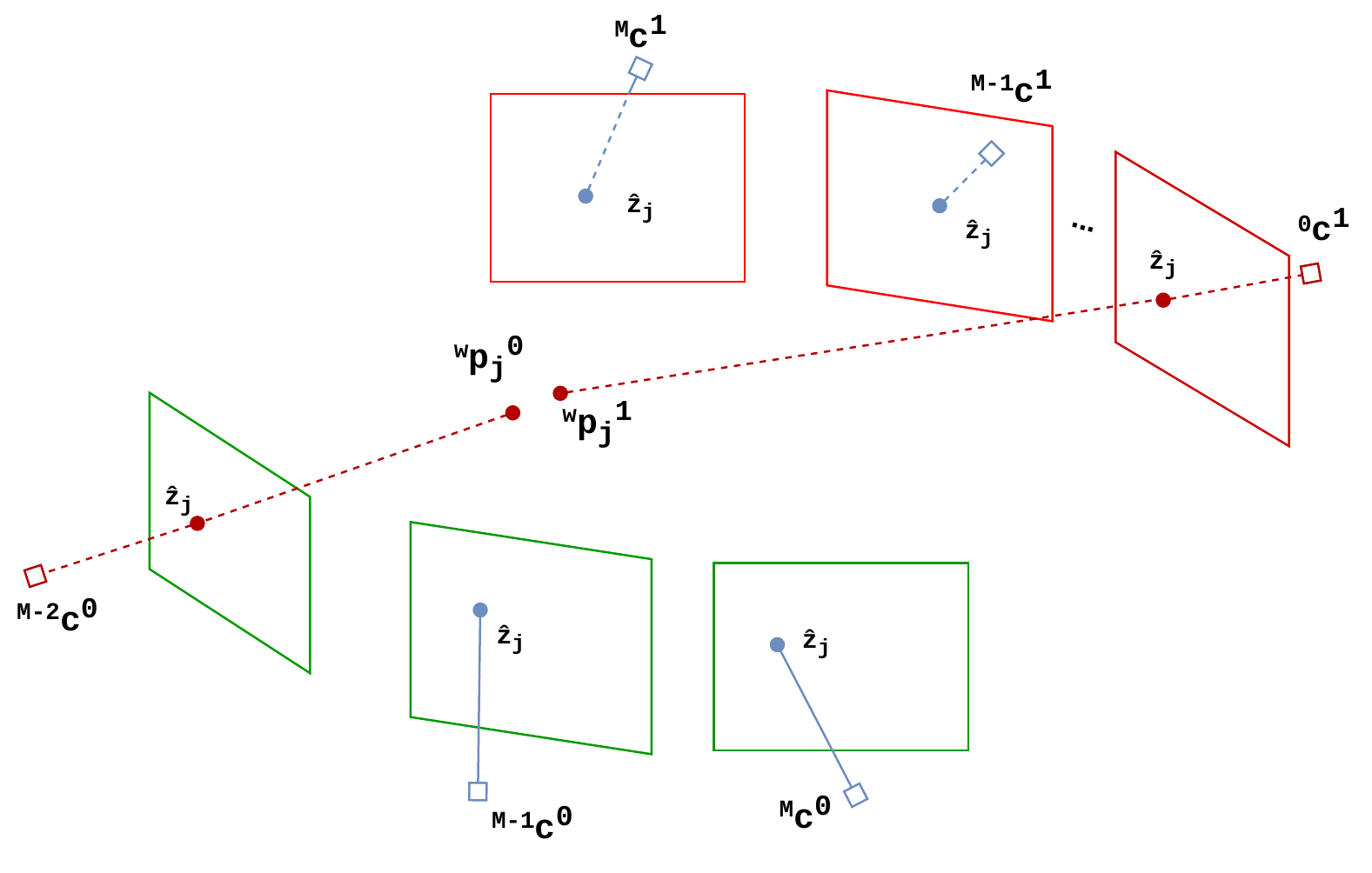}
    
    \caption{Collaborative SLAM representation. $^{M-2}c^0$ and $^0c^1$ are the anchor poses of two different agents $0$ (green) and $1$ (red) respectively. These anchors observed the landmark $\wpf$ first with respect to the other frames, $\wpf$ is parametrized according to $^{M-2}c^0$ for UAV $0$ and to $^0c^1$ for UAV $1$.}
    \label{fig:collab_slam}
    \vspace{-2ex}
\end{figure}

To perform an EKF measurement update with the resulting equation \eqref{eq:collab_msckf}, we need to multiply on each side by the left nullspace ${\boldsymbol{A}_{j}}$ of $\jacob{j}{\boldsymbol{p}}{}$ and by doing so obtain two subsystems as in  \eqref{eq:nullspace_sys}. Finally, we take the part that depends on the state only, 
\begin{equation}
\deltaresidualbarp{}{j}{}
= 
\jacob{j}{\state}{}
\deltastate +
\boldsymbol{n}_{j}^{}
\label{eq:collab_msckf_nullspace}
\end{equation}
and perform the CI-EKF \eqref{eq:ci-ekf-s}-\eqref{eq:ci-ekf-state} update to take into account the possible correlations between the agents states.

The Jacobian $\jacob{j}{\state}{}$ can be rewritten according to the state the various columns refer to as:
\begin{equation}
\deltaresidualbarp{}{j}{}
= 
\begin{bmatrix}
\jacob{}{}{l_0}
& \hdots & 
\jacob{}{}{l_L}
\end{bmatrix}
\deltastate +
\boldsymbol{n}_{j}^{}
\label{eq:collab_msckf_nullspace1}
\end{equation}

\subsection{SLAM-SLAM update}
To perform the collaborative update between two SLAM features, we adopted the approach proposed by Zhu~\cite{Zhu21iros} for its efficiency. The method constrains the EKF, that is, when two SLAM features of two different agents $0$ and $1$  match, it  implies that they refer to the same 3D point in the world frame and so the difference between the landmarks in the world frame must be zero:
\begin{equation}
    \wpfp{0}-\wpfp{1}=0
    \label{eq:slam_constraint}
\end{equation}
as \figref{fig:collab_slam} shows.

In \emph{xVIO} the SLAM landmarks are stored in the EKF state in terms of an inverse depth parametrization. The parametrization is calculated with respect to the anchor pose $\pawp{l}$, also part of the state, that is the pose of the camera frame that observed the feature point first. Hence, we can write \eqref{eq:slam_constraint} in terms of the states of the two robots:
\begin{equation}
\pawp{0} + \frac{1}{\rho_j^{0}}\Cawp{0}^T\begin{bmatrix} \alpha_j^{0} \\ \beta_j^{0} \\1 \end{bmatrix} -
\pawp{1} - \frac{1}{\rho_j^{1}}\Cawp{1}^T\begin{bmatrix} \alpha_j^{1} \\ \beta_j^{1} \\1 \end{bmatrix}  = 0
\label{eq:collab_slam}
\end{equation}
where $\alpha_j^{l}$, $\beta_j^{l}$, and $\rho_j^{l}$ are the inverse depth parametrization of the landmark $\wpfp{l}$ observed by the drone $l$. Linearizing \eqref{eq:collab_slam}, we can write the residual obtaining a model that depends on the states of the two UAVs involved in the update as in \eqref{eq:collab_msckf_nullspace1}, that is then used to perform the CI-EKF \eqref{eq:ci-ekf-s}-\eqref{eq:ci-ekf-state} update.
\changes{\subsection{EKF prediction step}}
\changes{The prediction step is the same as in \emph{xVIO}, that is the IMU measurements are propagated using the inertial integration model described in~\cite{Weiss11icra}. For sake of completeness we report the IMU measurement model but we refer the reader to Delaune et al.~\cite{delaune20xvio} work for more details:
\begin{align}
\boldsymbol{\omega}_{IMU}& = {^i\boldsymbol{\omega}_w^i} +\boldsymbol{b}_g +
	\boldsymbol{n}_g
\label{eq:imu-gyro-measurement}\\
\boldsymbol{a}_{IMU}&=\boldsymbol{C}(\boldsymbol{q}{}_w^i)
	(\boldsymbol{a}_w^i - ^w\boldsymbol{g}) + \boldsymbol{b}_a
	+ \boldsymbol{n}_a
\label{eq:imu-accelero-measurement}
\end{align}
where $n_g$ and $n_a$ are zero-mean Gaussian white noises, the biases $\boldsymbol{b}_g$ and $\boldsymbol{b}_a$ are modeled as random walk with zero-mean Gaussian white noise. The gravity vector $g$ is considered constant, the Coriolis forces and the planet curvature are neglected.
}

\changes{\subsection{EKF measurements update step}}
To fuse the data coming from different agents, we should keep track of the cross-correlation between the various states. \changes{However, doing so is expensive in terms of memory since the number of elements in each drone's covariance will increase quadratically with the number of agents sharing data.} The Covariance-Intersection (CI) algorithm~\cite{609105} allows us to neglect the correlation information between the robots, and perform data fusion with consistent covariance information. Namely, the resulting covariance out of the CI algorithm is larger than the real estimate one could obtain considering the cross-correlation. \changes{By doing so, the EKF is not overconfident, and it is kept consistent.} The CI algorithm consists of a convex minimization problem that finds the best weights $\omega_l$ to scale the covariances of each agent $l \in \{l_0,...l_L\}$ involved in the update to obtain a covariance with the minimum eigen values. 
We use the CI-EKF to update the drone state in the same form Zhu proposed, where $l_0$ denotes the drone where the update is performed:
\begin{eqnarray}
\label{eq:ci-ekf-s}
\boldsymbol{S} &=& \boldsymbol{R} + \sum_{l \in \{l_0,...l_L\}} \frac{1}{\omega_l}\jacob{}{}{l^T}\boldsymbol{P}_{k-1}^l\jacob{}{}{l}\\
\boldsymbol{K} &=& \frac{1}{\omega_{l_0}}\boldsymbol{P}_{k-1}^{l_0}\boldsymbol{H}^{{l_0}^T}\boldsymbol{S}^{-1} \\
\boldsymbol{P}_{k}^{l_0} &=& (\changes{\boldsymbol{I}_{\omega_{l_0}}}-\boldsymbol{K}\jacob{}{}{{l_0}^T})\frac{1}{\omega_{l_0}}\boldsymbol{P}_{k-1}^{l_0}\\
\deltastate^{l_0} &=& \changes{\boldsymbol{K}}\deltaresidualp{}{}{l_0}
\label{eq:ci-ekf-state}
\end{eqnarray}
Where $\boldsymbol{R}$ is the covariance matrix of the Gaussian noise described in Section \ref{sec:state-definition}. \changes{$\boldsymbol{P}_{k-1}^{l}$ is the covariance term of the drone $l$, whereas $\boldsymbol{P}_{k}^{l_0}$ is the new covariance of the agent that is performing the update. Notice that $\boldsymbol{I}_{\omega_{l_0}}$ is a diagonal matrix where each element $a_{i}^{\omega_{l_0}}$ of the diagonal is defined as:
\begin{equation}
    a_{i}^{\omega_{l_0}} =
    \begin{cases}
      1 & \mbox{if } i\in \mathbb{J} \\
      \omega_{l_0} & \mbox{otherwise}
    \end{cases}
\end{equation}
$\mathbb{J}$ is the set of ids in the state vector $\state$ that are directly involved in the update (i.e., the parts of the state that will be updated). This formulation is crucial to perform the CI update without making the EKF underconfident on the part of the state not involved in the collaborative update.
}

Before propagating the state and the covariance to remove outliers, the innovation of each collaborative update undergoes a $\chi^2$ test with $95\%$ confidence. We compute the Mahalanobis distance with the real covariance of each agent and not with the one scaled by the CI algorithm. The reasons behind this choice are twofold:
\begin{itemize}
    \item Being conservative: the original covariance is smaller than the scaled one and so the resulting distance is larger, hance is more difficult to pass the $\chi^2$ test.
    \item Avoid unuseful computation: we compute the CI covariance only if the test is passed.
\end{itemize}

\subsection{Communication Pipeline}
To perform the collaborative MSCKF and SLAM updates, the agents need the tracks, the state, and the covariance of the other robots. We define a message called \emph{MessageUAV} which contains all the information needed to perform the collaborative updates.
A naive approach to share data between agents would be one in which each robot sends a \emph{MessageUAV} to all the others at the camera frame rate. The result would be an extremely high usage of the communication channel considering that an estimate of the message size is around $190.7 kB$. We computed this estimate by averaging the weights of the \emph{MessageUAV} sent by the robots in simulation.

To avoid this significant usage of the bandwidth, we developed a keyframe-based request/response data exchange system to lower the data sent over the network. To perform the collaborative updates, the agents need to find correspondences between the features they are tracking and the tracks they receive. Hence, this is the same as performing place recognition and then loop-closure among a series of keyframes. 

Each drone sends to all the others a message named \emph{RequestUAV} at the camera frame rate. This message contains a Binary VLAD descriptor of the frame the agent is viewing. We use a modified version of the BVLAD~\cite{7025567}, to perform place recognition between the UAVs. The ORB descriptor is binary and so we create a VLAD by using binary operations only, in particular, we use the OR ($\vee$) and the AND ($\wedge$) operations, instead of sums and subtractions. Doing so we do not need to perform normalization nor exponential decay law to reduce the impact of recurring features. VLADs are of a fixed length, and they can be shortened applying dimensionality reduction through PCA~\cite{6619051}. The resulting request message has a fixed size that is much smaller than a \emph{MessageUAV}. We created a coarse vocabulary of ORB visual words of 64 centroids. Hence, without applying any dimensionality reduction sharing a VLAD is the same as sharing 64 ORB descriptors. The result is that the \emph{RequestUAV} message weighs $2.05 kB$, including the timestamp and the id of the sender, resulting in lower data consumption. 
When a robot receives the request message from another agent, it looks among its keyframes to find a loop-closure. A keyframe contains the information needed by the \emph{MessageUAV} and a VLAD. The score between the received descriptor $r_j$ and the keyframe one $r_k$ is given by:
\begin{equation}
    s \doteq \frac{Hamming(r_j \wedge r_k)}{D_{max}}
\end{equation}
Where $D_{max}$ is the maximum distance that can be computed between two descriptors and $Hamming(\cdot)$ is the Hamming distance of the resulting vector. If the score is over a certain threshold, the keyframe is wrapped into a message and shipped to the agent who sent the request.

Each agent maintains a database of keyframes. To populate the database, we follow the same approach proposed for PTAM~\cite{klein07parallel}. Namely, every time the ratio between the baseline $b$, that is the distance between the previous keyframe and the current camera pose, and the average depth $\bar{z}$ of the tracked features is higher than a threshold that is usually around $10-20\%$, a new keyframe is defined.

\section{Experiments}
\label{sec:experim}

\begin{figure}[b!]
    \centering
    \includegraphics[width=\linewidth, trim=8.5 8.0 8.0 8.0,clip]{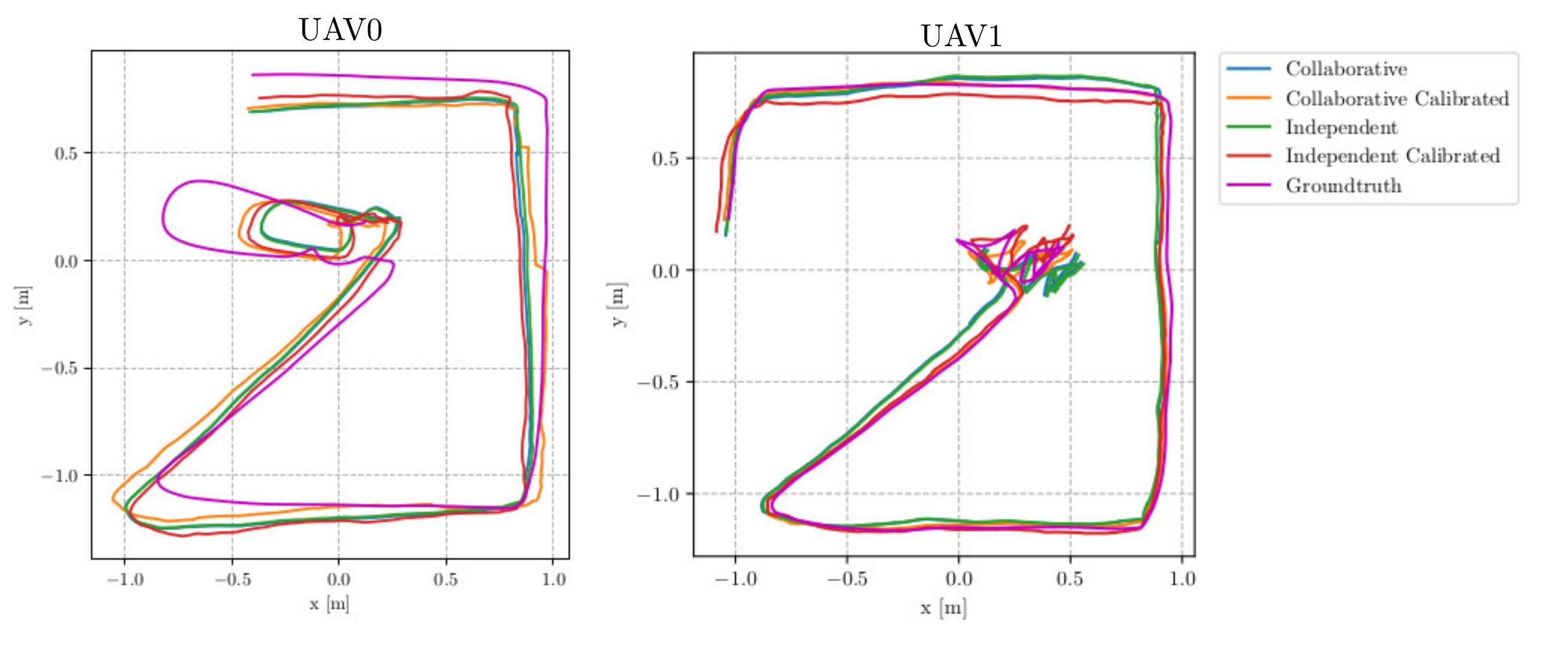}
    \vspace{-3ex}
    \caption{Trajectories estimate results for UAV0 and UAV1 in the Mars Yard dataset.}
   \label{fig:thermal_traj}
    \vspace{-2ex}
\end{figure}

\begin{figure*}[t!]
    \centering
    \includegraphics[width=0.33\linewidth, trim=0.2 0.0 0.2 0.0, clip]{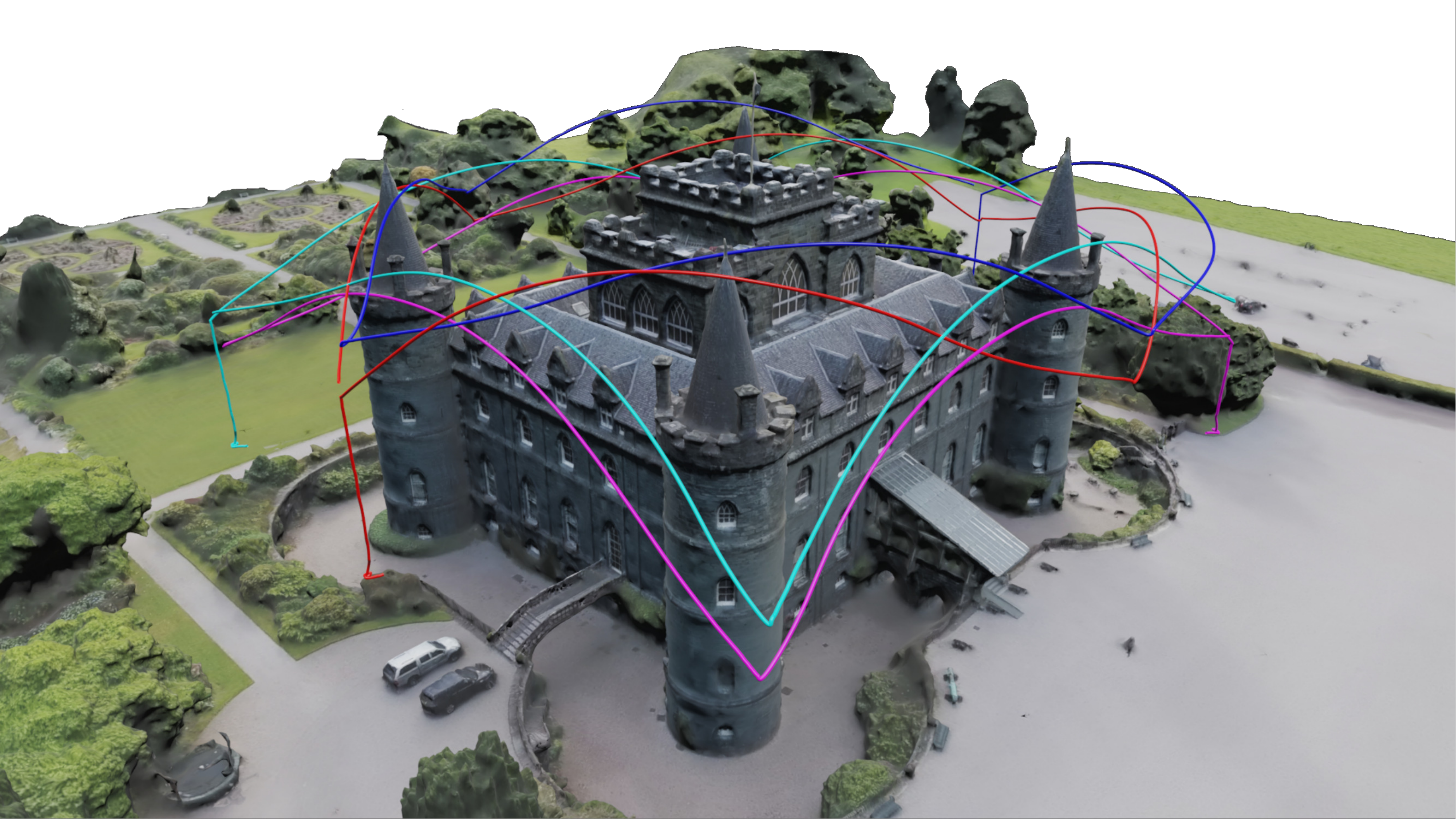}\hfill
    \includegraphics[width=0.33\linewidth, trim=0.2 0.0 0.2 0.0, clip]{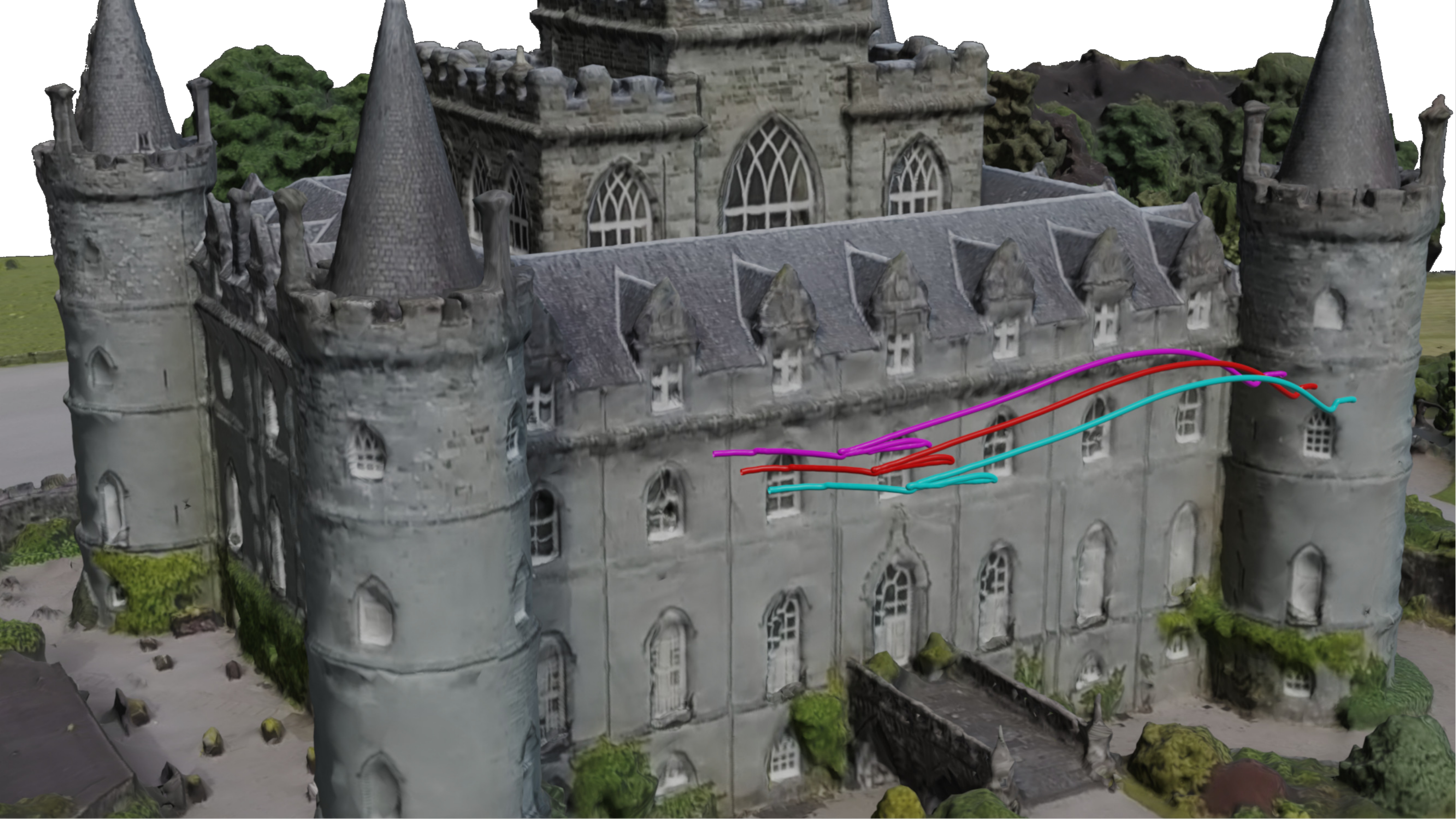}\hfill
    \includegraphics[width=0.33\linewidth, trim=1.0 1.0 1.0 1.0, clip]{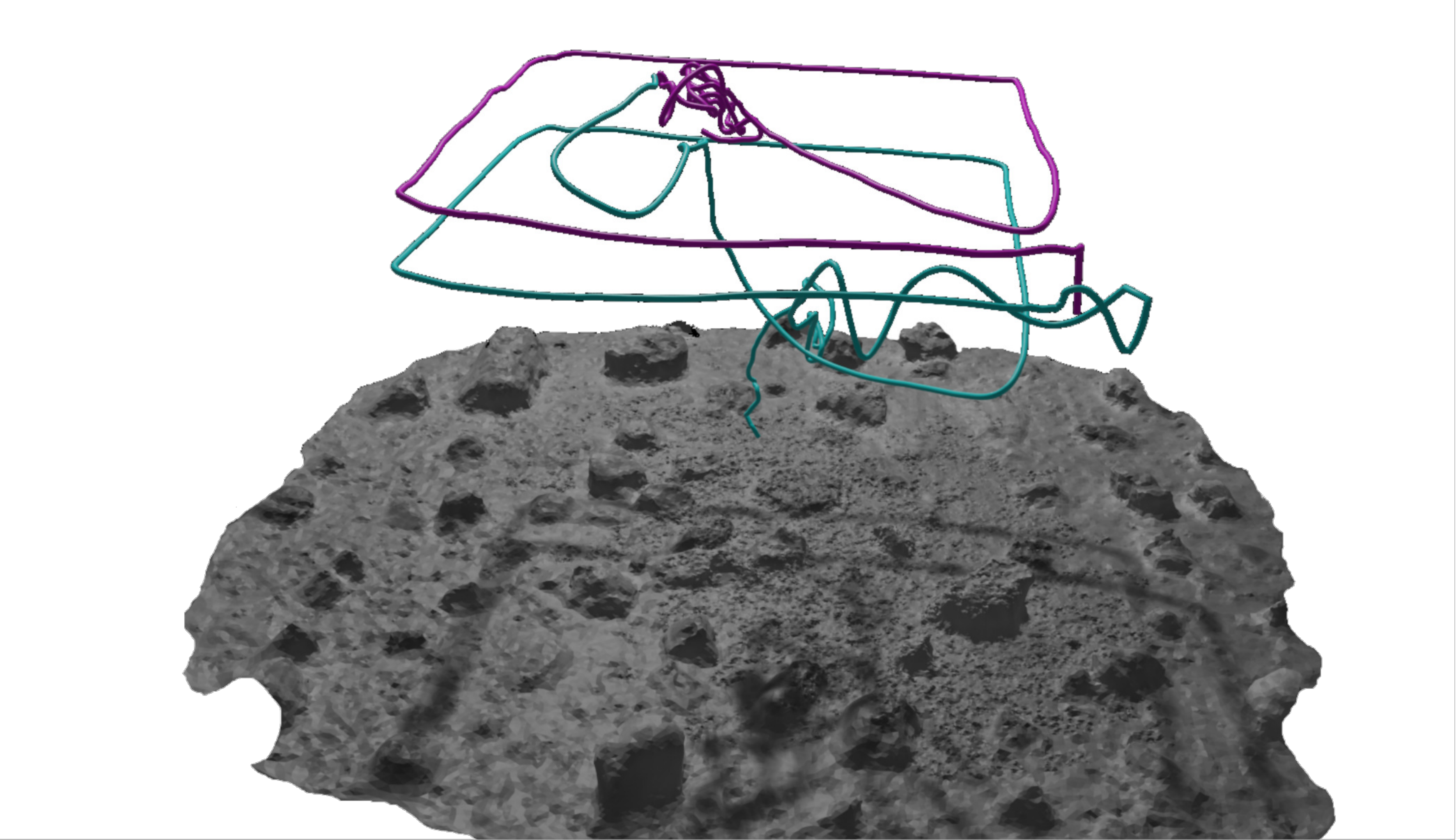}\hfill
    \caption{Dataset scenarios (a) "Castle Around" four drones fly around the
    Inveraray Castle model (https://skfb.ly/6z7Rr) by Andrea Spognetta licensed
    under Creative Commons Attribution-NonCommercial. (b) "Castle Parallel"
    three drones fly parallel trajectories in front of the castle. (c) \changes{Two
    drones fly squared trajectories on the Mars Yard at JPL.}
    }
    \label{fig:datasets}
\vspace{-2ex}
\end{figure*}

\begin{table}[tb]
\centering
\caption{Absolute trajectory error and standard deviation for the visual datasets.}
\begin{adjustbox}{max width=\linewidth}
\setlength{\tabcolsep}{16pt}
\begin{tabular}{llccc}
    & & Collaborative & xVIO~\cite{delaune20xvio} & Improvement\\
    \textbf{Dataset} & \textbf{Agent} & \textbf{Ours}  & baseline &\\ 
    \midrule 
    
    Castle & UAV0 & \textbf{0.4312$\pm$0.2632} & 0.6183$\pm$0.4136 & 30.26\% \\
    Parallel & UAV1 & \textbf{0.3871$\pm$0.1860} & 0.4817$\pm$0.3039 & 19.64\% \\
    & UAV2 & 0.6108$\pm$0.3730 & \textbf{0.5169$\pm$0.3153} & -18.17\% \\
    \midrule 
    & UAV0 & \textbf{0.3437$\pm$0.1678} & 0.3788$\pm$0.1973 & 9.26\% \\
    Castle & UAV1 & \textbf{0.3663$\pm$0.1531} & 0.7186$\pm$0.3547 & 49.02\% \\
    Around & UAV2 & \textbf{0.5670$\pm$0.2851} & 0.5995$\pm$0.3276 & 5.42\% \\
    & UAV3 & 0.8717$\pm$0.4094 & \textbf{0.7240$\pm$0.3524} & -20.40\% \\
    
\midrule
\end{tabular}
\end{adjustbox}
\label{tab:compare:visual_results}
\vspace{1ex}
\end{table}

\begin{table}[tb]
\centering
\caption{Absolute trajectory error and standard deviation for thermal Mars Yard dataset.}
\begin{adjustbox}{max width=\linewidth}
\setlength{\tabcolsep}{6pt}
\begin{tabular}{llcccc}
    & & Collaborative & Collaborative & Independent & xVIO~\cite{delaune20xvio}\\
    \textbf{Dataset} & \textbf{Agent} & Calibrated~\textbf{Ours} & \textbf{Ours} & Calibrated~\textbf{Ours} & Baseline \\ 
    \midrule 
    Mars & UAV0 & \textbf{0.1610$\pm$0.0731} & 0.1860$\pm$0.0928 & 0.1906$\pm$0.0882 & 0.1835$\pm$0.0926 \\
    Yard & UAV1 & \textbf{0.0438$\pm$0.0189} & 0.0812$\pm$0.0381 & 0.0621$\pm$0.0248 & 0.0813$\pm$0.0382 \\
    
\midrule
\end{tabular}
\end{adjustbox}
\vspace{-1ex}
\label{tab:compare:thermal_results}
\vspace{1ex}
\end{table}

\begin{table}[tb]
{\centering
\caption{A \emph{MessageUAV} object weighs 190.7 kB while a request message occupies 2.05 kB.}
\begin{adjustbox}{max width=\linewidth}
\setlength{\tabcolsep}{6pt}
\begin{tabular}{llcccc}
    \textbf{Dataset} & \textbf{Input} & Naive [kB] & (\textbf{Ours}) [kB] & Improvement \\
    \midrule
    
    Castle & Message UAV & 639.60 & \textbf{264.30} & 57.60\% \\
    Around & Request UAV & - & \textbf{6.89} & \\
    \midrule
    
    Castle & Message UAV & 152.56 & \textbf{141.12} & 6.42\% \\
    Parallel & Request UAV & - & \textbf{1.64} &\\
    \midrule
    
    Mars & Message UAV & 234.37 & \textbf{22.88} & 89.17\% \\
    Yard & Request UAV & - & \textbf{2.51} & \\
\midrule
\end{tabular}
\end{adjustbox}
}
\hrulefill{A \emph{MessageUAV} object weighs 190.7 kB while a request message occupies 2.05 kB. The data reported by the table are estimated according to the number of messages the agents exchange in simulation \changes{and real data scenarios}. The camera frame rate for all the datasets is 30Hz. The length in seconds for each dataset is 30s, 9s, 20s for the \textbf{Castle Around}, \textbf{Castle Parallel}, \textbf{Mars Yard} respectively.}
\vspace{-1ex}
\label{tab:compare_data_usage}
\vspace{1ex}
\end{table}

\begin{table}[tb]
\centering
\caption{Comparison between COVINS~\cite{Schmuck21ismar} and \changes{our approach}.}
\begin{adjustbox}{max width=\linewidth}
\setlength{\tabcolsep}{6pt}
\begin{tabular}{lccccc}
    \textbf{Algorithm} & \textbf{Agent} & \textbf{N. of Messages sent} & \textbf{Max CPU [\%]} & \textbf{Max Mem [MiB]} & \textbf{ATE [m]} \\
    \midrule
    
     & UAV0 & 4587 & 208.47 & 710.8 & \textbf{0.1641$\pm$0.0624} \\
    COVINS~\cite{Schmuck21ismar} & UAV1 & 4524 & 211.28 & 643.5 & \textbf{0.3626$\pm$0.1214} \\
     & UAV2 & 4457 & 204.87 & 681.1 & \textbf{0.2058$\pm$0.0614} \\
     & UAV3 & 4788 & 201.18 & 701.8 & \textbf{0.2897$\pm$0.1132} \\
     & Server & - & 325.78 & 710.9 & - \\
     \midrule
     
     & UAV0 & \textbf{1482} & \textbf{165.66} & \textbf{133.9} & 0.3437$\pm$0.1678 \\
    \textbf{Ours} & UAV1 & \textbf{1414} & \textbf{158.56} & \textbf{124.8} & 0.3663$\pm$0.1531 \\
     & UAV2 & \textbf{1475} & \textbf{154.18} & \textbf{139.1} & 0.5670$\pm$0.2851\\
     & UAV3 & \textbf{1465} & \textbf{170.79} & \textbf{136.2} & 0.8717$\pm$0.4094\\

\midrule
\end{tabular}
\end{adjustbox}
\vspace{-1ex}
\label{tab:covins_comparison}
\vspace{-2ex}
\end{table}

\figref{fig:datasets} shows the 3D representation of the environments we recorded the data for testing our system.
To validate our system, we used real thermal recordings from the JPL Mars Yard. The data was recorded from the industrial version of the FLIR Boson camera with a lens providing a 95-deg field of view (FOV). It generates 640 x 512 images in the $7.5 {\mu}m$ - $13.5 {\mu}m$  longwave infrared spectral range, with a thermal sensitivity under $40 mK$ and an 8-ms time constant. The camera was calibrated using as a target a circle grid pattern laminated over a Gatorfoam board left for 5 minutes in the sun, then the Kalibr toolbox~\cite{Furgale13iros} was used to find the calibration parameters. The IMU used was the ICM20608 on-board the Pixhawk mini. This dataset is the same as the collection used for the work by Delaune et al.~\cite{Delaune19iros}. We merged two \changes{Robot Operating System (ROS)} bag files where two UAVs fly two squared trajectories with different altitudes, over the JPL Mars Yard for almost $10 m$, \figref{fig:thermal_traj} illustrates the two drones trajectories with their estimates. The data is collected with a five-minute distance in time. Therefore, \changes{we can rely on the assumption that} the thermal fingerprint of the objects does not \changes{drastically change over time}.
The results of our system compared against the baseline that is the standard xVIO implementation are reported in Table \ref{tab:compare:thermal_results}. We used the RPG trajectory evaluator tool \cite{Zhang18iros} to compute the Absolute Trajectory Error between the ground-truth and the estimates. The collaborative setup with photometric calibrated thermal images outperforms the standard xVIO by~\SI{12.26}{\percent} and~\SI{46.12}{\percent}.

We also created two synthetic datasets with visual data to stress different aspects of our method. We modified the VI-Sensor Simulator~\cite{Teixeira20RAL} to generate ground-truth visual, inertial and landmark data to test the back-end of the system. The drones simulated by the VI-Sensor Simulator mount a visual camera that generates 752 x 480 images with a focal length of 455 px at 30 Hz. The IMU sensor simulated is the ADIS16448 that generates data at 200 Hz.
The datasets \emph{Castle Around} and \emph{Castle Parallel} are recorded using a 3D reconstruction of the Scottish Inveraray Castle. The former simulates four UAVs flying $220 m$ squared trajectories around the castle with different altitudes and rotations. The latter consists of 3 drones flying $30m$ parallel trajectories at the same altitude which stresses the communication pipeline. Table \ref{tab:compare:visual_results} shows the results using the collaborative approach. 
\\
Table \ref{tab:compare_data_usage} reports the different data usage between the proposed communication pipeline and the naive approach in the different datasets. Notably, our approach shows a decrease of data exchanged by~\SI{89}{\percent} in one of the scenarios. In the Castle Parallel the improvement is lower due to the trajectory design, where the drones observe the same scene and loop closure is performed all along the path. To simulate the different drones we used containerization, where each drone is a Docker container and runs independently from the others. The docker base image we used to run the collaborative setup is a modified version of the \emph{dt-base-environment} of the Duckietown platform~\cite{paull17icra}.
\\
Finally, we also compared our system with visual data against the state-of-the-art Centralized Cooperative system COVINS~\cite{Schmuck21ismar}. The results, reported in Table \ref{tab:covins_comparison}, show that the centralized method outperforms our system in terms of accuracy of the trajectory estimation in almost all of the agents. This result was expected since the COVINS system performs loop closure detection and global pose and map optimization. In contrast, our filter-based approach shows lower memory and CPU usage as well as a much lower number of messages exchanged in the network compared to COVINS.

\changes{For completeness, we evaluated the time performance of our methods. Table \ref{tab:time_res} reports the average time in milliseconds needed by each block of the proposed system to process the received information. The total time needed to update the agent's state is about $53ms$. Hence our system can perform a collaborative state estimate with a frequency of $20 Hz$. The implemented pipeline leaves a margin of improvement considering more efficient multi-threading approaches. }

\section{Conclusion}
\label{sec:conclusion}
\begin{table}[tb]
\centering
\caption{In the first row the average time needed by each module expressed in [ms]. The second row shows the computational time percentage occupied by each block. }
\begin{adjustbox}{max width=\linewidth}
\setlength{\tabcolsep}{6pt}
\begin{tabular}{cccccccc}
VLAD &    Feature    & Loop    & Photometric & Tracker & MultiSLAM & MultiMSCKF \& & \\
Generation & Matching & closure & calibration & & &  MSCKF & \textbf{Tot}\\
\midrule

    0.08 & 7.19 & 3.03 & 22.53 & 13.45 & 2.46 & 1.15 + 3.18  & 53.07 \\
    ~0\% & 13.55\% & 5.71\% & 42.44\% & 25.34\% & 4.63\% & 8.16\% & 100\% \\
\midrule
\label{tab:time_res}
\end{tabular}
\end{adjustbox}
\vspace{-4ex}
\end{table}
With this work, we extended the JPL xVIO library with a decentralized multi UAV
system that can work with thermal and visual data. To the best of our knowledge,
the described project is the first work that implements a collaborative
Thermal-Inertial Odometry system using a photometric calibration algorithm that
improves feature matching for loop closure detection. We developed a
multi-threading communication pipeline that reduces the bandwidth compared to a
naive approach and maintains system scalability. Our Code and associated datasets are open-source.

\section*{Acknowledgments}
The research was funded by the Combat Capabilities Development Command Soldier Center and Army Research Laboratory.
This research was carried out at the Jet Propulsion Laboratory, California
Institute of Technology, and was sponsored by the JPL Visiting Student Research
Program (JVSRP) and the National Aeronautics and Space Administration
(80NM0018D0004).

{\small
\balance
\bibliographystyle{IEEEtran}
\bibliography{all}

\begin{thebibliography}{10}
\providecommand{\url}[1]{#1}
\csname url@rmstyle\endcsname
\providecommand{\newblock}{\relax}
\providecommand{\bibinfo}[2]{#2}
\providecommand\BIBentrySTDinterwordspacing{\spaceskip=0pt\relax}
\providecommand\BIBentryALTinterwordstretchfactor{4}
\providecommand\BIBentryALTinterwordspacing{\spaceskip=\fontdimen2\font plus
\BIBentryALTinterwordstretchfactor\fontdimen3\font minus
  \fontdimen4\font\relax}
\providecommand\BIBforeignlanguage[2]{{%
\expandafter\ifx\csname l@#1\endcsname\relax
\typeout{** WARNING: IEEEtran.bst: No hyphenation pattern has been}%
\typeout{** loaded for the language `#1'. Using the pattern for}%
\typeout{** the default language instead.}%
\else
\language=\csname l@#1\endcsname
\fi
#2}}

\bibitem{MarCO}
J.~Schoolcraft, A.~T. Klesh, and T.~Werne, ``Marco: interplanetary mission
  development on a cubesat scale.'' in \emph{14th International Conference on
  Space Operations}, vol. 2491, 05 2016, pp. 1--8.

\bibitem{CADRE}
\BIBentryALTinterwordspacing
A.~Pirrami and H.~Smith, ``Cooperative autonomous distributed robotic explorers
  (cadre),'' 2021. [Online]. Available:
  \url{https://www.nasa.gov/directorates/spacetech/game_changing_development/projects/CADRE}
\BIBentrySTDinterwordspacing

\bibitem{Zhang19Springer}
Z.~Zhang and D.~Scaramuzza, ``Visual-inertial odometry of aerial robots,''
  \emph{Encyclopedia of Robotics, Springer}, 2019.

\bibitem{Huang19icra}
G.~Huang, ``Visual-inertial navigation: A concise review,'' in \emph{{IEEE}
  Int. Conf. Robot. Autom. (ICRA)}, 2019.

\bibitem{bayard2019vision}
D.~S. Bayard, D.~T. Conway, R.~Brockers, J.~H. Delaune, L.~H. Matthies, H.~F.
  Grip, G.~B. Merewether, T.~L. Brown, and A.~M. San~Martin, ``Vision-based
  navigation for the {NASA} {Mars} helicopter,'' in \emph{AIAA Scitech 2019
  Forum}, 2019, p. 1411.

\bibitem{APUFFER}
E.~R. Boroson, R.~Hewitt, N.~Ayanian, and J.-P. de~la Croix, ``Inter-robot
  range measurements in pose graph optimization,'' in \emph{2020 IEEE/RSJ
  International Conference on Intelligent Robots and Systems (IROS)}, 2020, pp.
  4806--4813.

\bibitem{SurveyThermalCameras}
R.~Gade and T.~Moeslund, ``Thermal cameras and applications: A survey,''
  \emph{Machine Vision and Applications}, vol.~25, pp. 245--262, 01 2014.

\bibitem{delaune20xvio}
J.~Delaune, D.~S. Bayard, and R.~Brockers, ``xvio: A range-visual-inertial
  odometry framework,'' pp. 1--67, 2020.

\bibitem{Schmuck17icra}
P.~Schmuck and M.~Chli, ``Multi-{UAV} collaborative monocular {SLAM},'' in
  \emph{{IEEE} Int. Conf. Robot. Autom. (ICRA)}, 2017, pp. 3863--3870.

\bibitem{Schmuck21ismar}
P.~Schmuck, T.~Ziegler, M.~Karrer, J.~Perraudin, and M.~Chli, ``Covins:
  Visual-inertial slam for centralized collaboration,'' in \emph{2021 IEEE
  International Symposium on Mixed and Augmented Reality Adjunct
  (ISMAR-Adjunct)}, 2021, pp. 171--176.

\bibitem{DecentralisedSLAMlowband}
E.~Nettleton, S.~Thrun, H.~Durrant-Whyte, and S.~Sukkarieh, ``Decentralised
  slam with low-bandwidth communication for teams of vehicles,'' vol.~24, 01
  2003, pp. 179--188.

\bibitem{gautam12iciis}
A.~Gautam and S.~Mohan, ``A review of research in multi-robot systems,'' in
  \emph{2012 IEEE 7th International Conference on Industrial and Information
  Systems (ICIIS)}, 2012, pp. 1--5.

\bibitem{Dong15icra}
J.~Dong, E.~Nelson, V.~Indelman, N.~Michael, and F.~Dellaert, ``Distributed
  real-time cooperative localization and mapping using an uncertainty-aware
  expectation maximization approach,'' in \emph{{IEEE} Int. Conf. Robot. Autom.
  (ICRA)}, 2015, pp. 5807--5814.

\bibitem{Arandjelovic16cvpr}
R.~Arandjelovi\'c, P.~Gronat, A.~Torii, T.~Pajdla, and J.~Sivic, ``{NetVLAD}:
  {CNN} architecture for weakly supervised place recognition,'' in \emph{{IEEE}
  Conf. Comput. Vis. Pattern Recog. (CVPR)}, 2016, pp. 5297--5307.

\bibitem{Cieslewski18icra}
T.~Cieslewski, S.~Choudhary, and D.~Scaramuzza, ``Data-efficient decentralized
  visual {SLAM},'' \emph{{IEEE} Int. Conf. Robot. Autom. (ICRA)}, 2018.

\bibitem{ArandjelovicNetVLAD16}
R.~Arandjelovi\'c, P.~Gronat, A.~Torii, T.~Pajdla, and J.~Sivic, ``{NetVLAD}:
  {CNN} architecture for weakly supervised place recognition,'' in \emph{IEEE
  Conference on Computer Vision and Pattern Recognition}, 2016.

\bibitem{Strasdat12jivc}
H.~Strasdat, J.~Montiel, and A.~Davison, ``Visual {SLAM}: Why filter?''
  \emph{Image Vis. Comput.}, 2012.

\bibitem{Mourikis07icra}
A.~I. Mourikis and S.~I. Roumeliotis, ``A multi-state constraint {K}alman
  filter for vision-aided inertial navigation,'' in \emph{{IEEE} Int. Conf.
  Robot. Autom. (ICRA)}, 2007, pp. 3565--3572.

\bibitem{Li13ijrr}
M.~Li and A.~I. Mourikis, ``High-precision, consistent {EKF}-based
  visual-inertial odometry,'' \emph{Int. J. Robot. Research}, vol.~32, no.~6,
  pp. 690--711, 2013.

\bibitem{609105}
S.~Julier and J.~Uhlmann, ``A non-divergent estimation algorithm in the
  presence of unknown correlations,'' in \emph{Proceedings of the 1997 American
  Control Conference (Cat. No.97CH36041)}, vol.~4, 1997, pp. 2369--2373 vol.4.

\bibitem{Zhu21icra}
P.~Zhu, Y.~Yang, W.~Ren, and G.~Huang, ``Cooperative visual-inertial
  odometry,'' in \emph{2021 IEEE International Conference on Robotics and
  Automation (ICRA)}, 2021, pp. 13\,135--13\,141.

\bibitem{Zhu21iros}
P.~Zhu, P.~Geneva, W.~Ren, and G.~Huang, ``Distributed visual-inertial
  cooperative localization,'' in \emph{2021 IEEE/RSJ International Conference
  on Intelligent Robots and Systems (IROS)}, 2021, pp. 8714--8721.

\bibitem{Delaune19iros}
J.~Delaune, R.~Hewitt, L.~Lytle, C.~Sorice, R.~Thakker, and L.~Matthies,
  ``Thermal-inertial odometry for autonomous flight throughout the night,'' in
  \emph{2019 IEEE/RSJ International Conference on Intelligent Robots and
  Systems (IROS)}, 2019, pp. 1122--1128.

\bibitem{khattak19icra}
S.~Khattak, C.~Papachristos, and K.~Alexis, ``Keyframe-based direct
  thermal–inertial odometry,'' in \emph{2019 International Conference on
  Robotics and Automation (ICRA)}, 2019, pp. 3563--3569.

\bibitem{Das2021}
\BIBentryALTinterwordspacing
M.~P. Das, L.~Matthies, and S.~Daftry, ``Online photometric calibration of
  automatic gain thermal infrared cameras,'' p. 2453–2460, Apr 2021.
  [Online]. Available: \url{http://dx.doi.org/10.1109/LRA.2021.3061401}
\BIBentrySTDinterwordspacing

\bibitem{11744023_34}
E.~Rosten and T.~Drummond, ``Machine learning for high-speed corner
  detection,'' in \emph{Computer Vision -- ECCV 2006}, A.~Leonardis,
  H.~Bischof, and A.~Pinz, Eds.\hskip 1em plus 0.5em minus 0.4em\relax Berlin,
  Heidelberg: Springer Berlin Heidelberg, 2006, pp. 430--443.

\bibitem{Bouguet1999PyramidalIO}
J.-Y. Bouguet, ``Pyramidal implementation of the lucas kanade feature
  tracker.''\hskip 1em plus 0.5em minus 0.4em\relax Intel Corporation,
  Microprocessor Res. Labs, 1999, pp. 1--9.

\bibitem{6126544}
E.~Rublee, V.~Rabaud, K.~Konolige, and G.~Bradski, ``Orb: An efficient
  alternative to sift or surf,'' in \emph{2011 International Conference on
  Computer Vision}, 2011, pp. 2564--2571.

\bibitem{Weiss11icra}
S.~Weiss and R.~Siegwart, ``Real-time metric state estimation for modular
  vision-inertial systems,'' in \emph{2011 IEEE International Conference on
  Robotics and Automation}, 2011, pp. 4531--4537.

\bibitem{7025567}
D.~Van~Opdenbosch, G.~Schroth, R.~Huitl, S.~Hilsenbeck, A.~Garcea, and
  E.~Steinbach, ``Camera-based indoor positioning using scalable streaming of
  compressed binary image signatures,'' in \emph{2014 IEEE International
  Conference on Image Processing (ICIP)}, 2014, pp. 2804--2808.

\bibitem{6619051}
R.~Arandjelovic and A.~Zisserman, ``All about vlad,'' in \emph{2013 IEEE
  Conference on Computer Vision and Pattern Recognition}, 2013, pp. 1578--1585.

\bibitem{klein07parallel}
G.~Klein and D.~Murray, ``Parallel tracking and mapping for small {AR}
  workspaces,'' in \emph{Proc. Sixth {IEEE} and {ACM} International Symposium
  on Mixed and Augmented Reality {(ISMAR'07)}}, Nara, Japan, November 2007.

\bibitem{Furgale13iros}
P.~Furgale, J.~Rehder, and R.~Siegwart, ``Unified temporal and spatial
  calibration for multi-sensor systems,'' in \emph{IEEE/RSJ Int. Conf. Intell.
  Robot. Syst. (IROS)}, 2013.

\bibitem{Zhang18iros}
Z.~Zhang and D.~Scaramuzza, ``A tutorial on quantitative trajectory evaluation
  for visual(-inertial) odometry,'' in \emph{IEEE/RSJ Int. Conf. Intell. Robot.
  Syst. (IROS)}, 2018.

\bibitem{Teixeira20RAL}
L.~Teixeira, M.~R. Oswald, M.~Pollefeys, and M.~Chli, ``{Aerial Single-View
  Depth Completion with Image-Guided Uncertainty Estimation},'' \emph{{IEEE}
  Robotics and Automation Letters ({RA-L})}, 2020.

\bibitem{paull17icra}
L.~Paull, J.~Tani, H.~Ahn, J.~Alonso-Mora, L.~Carlone, M.~Cap, Y.~F. Chen,
  C.~Choi, J.~Dusek, Y.~Fang, D.~Hoehener, S.-Y. Liu, M.~Novitzky, I.~F.
  Okuyama, J.~Pazis, G.~Rosman, V.~Varricchio, H.-C. Wang, D.~Yershov, H.~Zhao,
  M.~Benjamin, C.~Carr, M.~Zuber, S.~Karaman, E.~Frazzoli, D.~Del~Vecchio,
  D.~Rus, J.~How, J.~Leonard, and A.~Censi, ``Duckietown: An open, inexpensive
  and flexible platform for autonomy education and research,'' in \emph{2017
  IEEE International Conference on Robotics and Automation (ICRA)}, 2017, pp.
  1497--1504.

\end{thebibliography}
}

\end{document}